\documentclass[letterpaper]{article}
\usepackage[T1]{fontenc}
\usepackage{uai2016}
\usepackage[margin=1in]{geometry}

\usepackage{amsmath}
\usepackage{syntax}
\usepackage{booktabs}
\usepackage{mdwlist}
\usepackage{hyphenat}

\usepackage{url}

\usepackage{fancyvrb}
\fvset{commandchars=\\\{\},hfuzz=2.5pt}
\newcommand{\leadsfrom}{\char`\<\char`\~}

\usepackage{times}
\usepackage{graphicx} % more modern
\usepackage{subfigure}

\usepackage{natbib1}
\bibpunct[; ]{(}{)}{;}{a}{}{,}%

\usepackage{cleveref}
\crefname{appendix}{Appendix}{Appendices}
\crefname{section}{Section}{Sections}
\crefname{figure}{Figure}{Figures}
\crefname{table}{Table}{Tables}

\crefname{algocf}{Algorithm}{Algorithms}

\usepackage[algo2e,ruled]{algorithm2e}


\long\def\remark#1{% for notes in the margin -- from Norman Ramsey
    \ifvmode\else
        \unskip\raisebox{-3.5pt}[0pt][0pt]{\rlap{$\scriptstyle\diamond$}}%
    \fi
    \marginpar{\raggedright\hbadness=10000
    \parindent=8pt \parskip=2pt
    \def\baselinestretch{0.8}\tiny
    \itshape\noindent #1\par}}
\usepackage{comment}

\title{Composing inference algorithms as program transformations}
\author{ {\bf Robert Zinkov} \\
         Indiana University \\
         Bloomington, IN 47408 USA \\
         \texttt{zinkov@iu.edu} \\
         \And
         {\bf Chung-chieh Shan} \\
         Indiana University \\
         Bloomington, IN 47408 USA \\
         \texttt{ccshan@indiana.edu}
       }

\begin{document}

\maketitle

\begin{abstract}

Probabilistic inference procedures are usually coded painstakingly from
scratch, for each target model and each inference algorithm.  We reduce
this effort by generating inference procedures from models
automatically.  We make this code generation modular by decomposing
inference algorithms into reusable program-to-program transformations.  These
transformations perform exact inference as well as
generate probabilistic programs that compute expectations, densities,
and MCMC samples.  The resulting inference procedures are about as
accurate and fast as other probabilistic programming systems on
real-world problems.

\end{abstract}

\section{INTRODUCTION}

Writing inference algorithms for probabilistic models is tedious and
error-prone. Conceptually, these algorithms are combinations
of simpler operations, such as computing the density of a distribution
at a given point. So it is unfortunate that these algorithms are traditionally
implemented from scratch. In this paper, we show how to describe these
building blocks in code, so that they need not be rewritten
for every new inference algorithm or model.

We contribute the first method for composing multiple
inference algorithms over the same model, even exact and approximate
ones over the same factor.  Our approach is to express inference in
terms of operations that transform one
probabilistic program into another.
We use probabilistic programs to represent distributions, though our
approach is compatible with other representations such as factor graphs.
The goal of our transformations
is to turn a program that denotes a model into another
program that, when interpreted to draw a weighted sample,
is equivalent to the desired inference algorithm.

Because the output of an inference transformation is still
a probabilistic program, we can apply further inference transformations
to the program.  In this way, we can subject a single model to multiple
inference methods without coding them from scratch.  We thus reduce the
informal problem of combining inference methods to the formal and more
automatable problem of composing program transformations.  In particular,
approximate inference methods can be composed with taking advantage of
exact mathematical equivalences such as conjugacy.
\begin{comment}
Chad: To be really picky, it sounds like you're not really subjective one model to multiple inference methods, but rather subjecting the result of one inference algorithm to another, chaining them together sequentially.
\end{comment}

\section{MOTIVATION AND RELATED WORK}

Developing inference algorithms that work on a variety of models has
long been a goal of probabilistic inference, including graphical
models and probabilistic programming.  The composability of inference
algorithms has unfortunately lagged behind the composability of models.

Many probabilistic programming systems allow a choice of
inference methods, both exact and approximate.
For example, the probabilistic language Church \citep{GMR+08} has many
interpreters, each of which implements a different inference method.
Systems such as Figaro, Factorie, Anglican and Wolfe
\citep{pfeffer2009figaro,MRWSS08,wood-aistats-2014,RSSRVN2014} also
allow adding inference methods, as new code in the
host language where the systems are embedded.  However, the end result of
applying these inference methods is behavior or code in a different
language, no longer a probabilistic program.  Thus, it is difficult in
these systems to apply one method to the
result of another method.

More similar to our approach is \citets[\citeyear{scibiormodular}]{DBLP:conf/haskell/ScibiorGG15}.
Like us, they express and compose inference methods as transformations
that produce probabilistic programs in the same language.  Thus for
example they reuse Sequential Monte Carlo to implement Particle
Independent \nolinebreak[3] MH\@.  But because their probabilistic language reuses many
primitives from the host language Haskell, their transformations cannot
inspect most of the input code, notably deterministic
computations and code under a binding.  In contrast, we
can perform exact inference (\cref{simpl}), we can compute densities and
conditional distributions (\cref{ssec:disintegrate}) in the face of
deterministic dependencies, and we can
generate MH samplers (\cref{s:mh}) using a variety of proposal distributions.

So in general, transformations need to take programs as input as well as
produce them as output in order to
support the variety of inference composition found in
the literature.
To illustrate the need and the variety, below we recall some patterns of
inference composition where we want to \emph{reuse} existing
implementations in ways unsupported by existing systems such as those
named above.

Sometimes, we apply approximate inference to a model,
then post-process the results using exact inference. For example, a
popular way to perform inference for latent Dirichlet allocation (LDA) is to
use Gibbs sampling \citep{griffiths2004finding} to infer topic markers
for each word, then infer from these topic markers the
exact distribution on words given each topic.

Other times, we apply exact inference to parts of our model, and use an
approximate method for the rest. As an example,
\citet{hughes2015reliable} develop an inference algorithm for
hierarchical Dirichlet processes that samples the truncation dictating the
number of topics then performs variational inference
for the other model parameters.
This inference combination
requires that sampling a truncation still leave in place a
representation on which we can perform variational inference.

Another composition pattern emerges from recent work
on parallelizing an inference algorithm to run on multiple
machines \citep{neiswanger2014asymptotically, NIPS20145596,
  2014arXiv1412.4869G}.
The pattern
is to transform a posterior distribution for a parameter given
the data into a model that lets us infer noisy versions of the parameter
given subsets of the data.
We then combine these noisy parameter estimates to infer the underlying
parameter.

Finally, given a linear sequential model, we often
want to predict future states of the system and the dynamics that
govern them. Given the dynamics, for systems like Kalman filters, we
may use exact inference to derive the state transition functions in
closed form. Learning the dynamics, on the other hand, is usually
treated as an approximate inference problem where we sample different
possible dynamics given some observed states. This joint learning and
exact inference again composes two
inference algorithms.
\Cref{easyroad} shows our inference
composition at work in a tiny instance of this case.
We illustrate our approach using this example in \cref{sec:example}.

\section{EXAMPLE OF INFERENCE COMPOSITION}
\label{sec:example}

We illustrate our program-transformation approach to inference
composition using a simple linear dynamical system.
Our model below defines a joint distribution:
\begin{alignat*}{3}
    &\mathrm{noise}_T  \hidewidth&& &&\sim \mathrm{Uniform}(3, 8)\\
    &\mathrm{noise}_E  \hidewidth&& &&\sim \mathrm{Uniform}(1, 4)\displaybreak[0]\\
    &x_1 &&\mathbin|      \mathrm{noise}_T &&\sim \mathrm{Normal}(0,   \mathrm{noise}_T) \\
    &m_1 &&\mathbin| x_1, \mathrm{noise}_E &&\sim \mathrm{Normal}(x_1, \mathrm{noise}_E) \displaybreak[0]\\
    &x_2 &&\mathbin| x_1, \mathrm{noise}_T &&\sim \mathrm{Normal}(x_1, \mathrm{noise}_T) \\
    &m_2 &&\mathbin| x_2, \mathrm{noise}_E &&\sim \mathrm{Normal}(x_2, \mathrm{noise}_E)
\end{alignat*}

We would like to draw samples from the posterior distribution
over $\mathrm{noise}_T$ and $\mathrm{noise}_E$
given observations $m_1$ and $m_2$,
using a Metropolis-Hastings (MH) sampler.

We start by representing the model in our language:

\begin{BVerbatim}
kalman =
 noiseT <~ Uniform(3, 8);
 noiseE <~ Uniform(1, 4);
 x1     <~ Normal( 0, noiseT);
 m1     <~ Normal(x1, noiseE);
 x2     <~ Normal(x1, noiseT);
 m2     <~ Normal(x2, noiseE);
 Dirac(((m1, m2), (noiseT, noiseE)))
\end{BVerbatim}

The use of \texttt{Dirac} at the bottom shows that this distribution
ranges over pairs of pairs of reals.

We first apply the \emph{disintegration} transformation to get another
program.  As detailed in \cref{ssec:disintegrate},
disintegration takes as input a joint
distribution and produces a program representing a family of
posterior distributions.
\begin{comment}
Chad: I see what you're saying, but only because I already know what it is. How about "Disintegration takes a joint distribution over a pair, $P(x,y)$, and computes the conditional distribution $P(y|x)$"
\end{comment}
The new program is a function
from the observations to the posterior distribution. For example,
disintegrating \texttt{kalman}
produces the program \texttt{kalman2} below.  It takes as
input \texttt{(m1, m2)} and returns the distribution over
\texttt{(noiseT, noiseE)} given those values for
\texttt{m1} and~\texttt{m2}.

\begin{BVerbatim}
kalman2 = Lam((m1,m2),
 noiseT <~ Uniform(3, 8);
 noiseE <~ Uniform(1, 4);
 x1     <~ Normal( 0, noiseT);
 x2     <~ Normal(x1, noiseT);
 Weight( exp(-(m2-x2)^2/(2*noiseE^2))
         /noiseE/sqrt(2*pi)
       * exp(-(m1-x1)^2/(2*noiseE^2))
         /noiseE/sqrt(2*pi)
       , (noiseT, noiseE) ))
\end{BVerbatim}

The use of \texttt{Lam} at the top and \texttt{Weight} at the bottom
shows that this is a function from pairs of reals \texttt{(m1, m2)} to
measures over pairs of reals \texttt{(noiseT, noiseE)}.

\begin{comment}
Chad: These new transformations come up, and the reader has no sense of how many more are to come, and how they all connect. Maybe some bullets about disintegrate, simplify, etc in the introduction would help this flow better
\end{comment}
The next step is to apply the \emph{simplification} transformation to
\texttt{kalman2} to get \texttt{kalman3}.

\begin{BVerbatim}
kalman3 = Lam((m1,m2),
 noiseT <~ Uniform(3, 8);
 noiseE <~ Uniform(1, 4);
 Weight(\(P\), (noiseT, noiseE)))
\end{BVerbatim}

This program is equivalent to \texttt{kalman2}, except the
simplification transformation has
symbolically
integrated out the \texttt{Normal}-distributed random variables
\texttt{x1} and~\texttt{x2} and replaced them by an observation
likelihood in closed form, which we elide above as~$P$.

We next apply to \texttt{kalman3} another program transformation we
call \emph{mh}, which implements MH sampling. The mh
transformation takes as input two programs. The first program represents
a proposal distribution, or more precisely a function from the current
sample to a distribution over proposed samples. Here we use
a proposal distribution that with equal probability resamples
one of \texttt{noiseT} and \texttt{noiseE} while keeping the other
fixed:

\begin{BVerbatim}
proposal = Lam((noiseT, noiseE),
 Superpose((1/2, n <~ Uniform(3, 8);
                 Dirac((n, noiseE))),
           (1/2, n <~ Uniform(1, 4);
                 Dirac((noiseT, n)))))
\end{BVerbatim}

The second input to the mh transformation represents the target
distribution. In this example, it is the part of \texttt{kalman3} above
after the top line \texttt{Lam((m1,m2),}.
From these inputs, the mh transformation computes a symbolic formula
for the MH acceptance ratio and embeds it in a program
representing a transition kernel.
The new program, \texttt{kalman4} below, is a function from the current sample to a distribution
over pairs of proposed samples and acceptance ratios:

\begin{BVerbatim}
kalman4 = Lam((noiseT, noiseE),
 proposed <~ Superpose(
   (1/2, n <~ Uniform(3, 8);
         Dirac((n, noiseE))),
   (1/2, n <~ Uniform(1, 4);
         Dirac((noiseT, n)))),
 Dirac((proposed, \(A\))))
\end{BVerbatim}

The elided part~$A$
is a symbolic formula that
computes the acceptance ratio using the current sample \texttt{(noiseT,
noiseE)} and the sample \texttt{proposed} by the \texttt{Superpose}.
This acceptance ratio can then be used to decide whether to accept or
reject the \texttt{proposed}.

We then perform further optimizations on \texttt{kalman4}, including
algebraic simplifications and rewriting the program to use fewer
\texttt{\leadsfrom}s. We
describe in more detail the kinds of optimizations we perform in
\cref{simpl}. The resulting program, \texttt{kalman5}, has the following
structure:

\begin{BVerbatim}
kalman5 = Lam((noiseT, noiseE),
 Superpose(
  (1/2, n <~ Uniform(3, 8);
        Dirac(((n, noiseE), \(A_T\)))),
  (1/2, n <~ Uniform(1, 4);
        Dirac(((noiseT, n), \(A_E\))))))
\end{BVerbatim}

The elided parts $A_T$ and~$A_E$
are algebraically simplified formulas for the acceptance
ratio in each of the two cases.

Finally we feed this last program \texttt{kalman5} to a sampler
(\cref{alg:is}).  Given an observation and a current
sample, this sampler produces a proposed sample and the
MH acceptance ratio of that sample.

\begin{BVerbatim}
sample(App(App(kalman5,(0,1)),(5,2)),
       [])
>>> (((5,1.6811397),0.7924639),1.0)
\end{BVerbatim}

In the command above, \texttt{(0,1)} is the observation, and
\texttt{(5,2)} is the current sample.  In the output above,
\texttt{(5,1.6811397)} is the proposed sample, and
\texttt{0.7924639} is its acceptance ratio.

\section{INFERENCE METHODS AS PROGRAM TRANSFORMATIONS}
\label{transforms}

To compose inference methods,
we pose them as transformations of one
probabilistic program into another. We then achieve the desired
inference method for the former program by applying a simpler inference
method, such as exact inference or weighted sampling, to the latter
program. For example, in \cref{sec:example} we feed a program to
disintegration (\cref{ssec:disintegrate}), then mh (\cref{s:mh}), then
simplification (\cref{simpl}), and finally sampling. Only in the final
sampling step is any random choice made!

We first define our probabilistic language,
then describe various program transformations
that work in concert.

\subsection{LANGUAGE DESCRIPTION}

Below is our core grammar of probabilistic programs:
\begin{grammar}
  <e> ::= <x> | "1" | <e> "-" <e> | <e> "<" <e> | "exp("<e>")" | "If("<e>","<e>","<e>")" | \dots
\alt "Sum("<e>","<e>","<x>","<e>")" | "Int("<e>","<e>","<x>","<e>")"
\alt "Lam("<x>","<e>")" | "App("<e>","<e>")" | "("<e>","<e>")" | <e>"[0]" | <e>"[1]"
\alt "Uniform("<e>","<e>")" | "Normal("<e>","<e>")"
\alt "Gamma("<e>","<e>")" | "Weight("<e>","<e>")"
\alt "Categorical(("<e>","<e>"), \dots)"
\alt "Superpose(("<e>","<e>"), \dots)" | <x>"\,<~\,"<e>"\,;\,"<e>
\end{grammar}
The first line of this grammar says that our language includes ordinary
programming support for variables, math,
and~\texttt{If}.
The second line adds primitives to represent \texttt{Sum}mation
and \texttt{Int}egration,
used in \cref{s:expectation}.
The third line adds functions and tuples.

%% We write
%% \texttt{Let(x,e1,e2)} as syntactic sugar for \texttt{App(Lam(x,e1),e2)}.

The remainder of the grammar is what makes our language probabilistic:
we add primitives that represent and compose measures. To start with,
\texttt{Uniform(1,2)} represents the uniform distribution
over real numbers between $1$ and~$2$, and \texttt{Normal(3,4)}
represents the normal distribution with mean~$3$ and standard
deviation~$4$.

\texttt{Weight(1,8)} represents the probability distribution that
assigns its entire probability mass~$1$ to the single outcome~$8$.
We write \texttt{Dirac(8)} as syntactic sugar for it.  In contrast,
\texttt{Weight(0.7,8)} represents the measure, or unnormalized
distribution, that assigns the probability~$0.7$ to the single
outcome~$8$. This primitive lets our language represent (unnormalized)
measures in general, not just (normalized) probability distributions.
This expressivity lets us separately reuse a transformation that produces an
unnormalized measure (\cref{ssec:disintegrate}) and a transformation
that subsequently normalizes a measure (\cref{ssec:normalization}).
Also, \texttt{Weight} lets us represent a distribution by combining
a base measure and a density
function.

\texttt{Categorical} represents the categorical
distribution with a sequence of zero or more pairs. The first element
of each pair is the probability of selecting the outcome that is the
second element of the pair. If the first elements
of the pairs do not sum to~1, they are normalized.

\texttt{Superpose} is like \texttt{Categorical}, except it does not
normalize, so it can represent measures in general.
We can define \texttt{Superpose} in
terms of \texttt{Categorical} and \texttt{Weight}, but it is actually
more convenient to
define \texttt{Weight} and \texttt{Categorical} in terms of
\texttt{Superpose}.

The final primitive \texttt{\leadsfrom} (pronounced ``bind'')
composes two distributions \texttt{e1}
and~\texttt{e2}.  The second distribution~\texttt{e2} may depend on the
outcome~\texttt{x} of~\texttt{e1}.  The outcome
of the composed distribution \texttt{x\leadsfrom e1;e2} is the outcome
of~\texttt{e2}. This primitive lets our language represent sequential
and hierarchical models. A simple example is this model:
\[
    x     \sim \mathrm{Uniform}(0, 2)\qquad
    y | x \sim \mathrm{Uniform}(x, 3)
\]
We write the marginal distribution over~$y$ as

\begin{BVerbatim}
x <~ Uniform(0, 2); Uniform(x, 3)
\end{BVerbatim}

and the joint distribution over $(x,y)$ as

\begin{BVerbatim}
x <~ Uniform(0, 2);
y <~ Uniform(x, 3); Dirac((x,y))
\end{BVerbatim}

% Thus, \texttt{Dirac} and \texttt{\leadsfrom} are the basic operations of the
% \emph{measure monad} \citep{giry-categorical,ramsey-stochastic}.

\begin{algorithm2e}[t]
  \KwIn{program representing a measure: $m$}
  \KwIn{environment: \textit{env}}
  \KwOut{pair of values (outcome, weight)}
  \textbf{Examine }$m$\\
  \uIf{$m$ has the form \texttt{Weight($w_1$,$e_1$)}}{ Evaluate $e_1$
      in the environment \textit{env},
      obtaining $v_1$ \\ Return $(v_1, w_1)$}
      \uElseIf{$m$ has the form \texttt{Normal($e_1$,$e_2$)}}{ Evaluate $e_1$ in the environment \textit{env},
      obtaining $v_1$ \\ Evaluate $e_2$ in the environment \textit{env},
      obtaining $v_2$ \\ Sample from the normal distribution
      with mean~$v_1$ and standard deviation~$v_2$, obtaining $v_3$ \\
      Return $(v_3, 1)$}
     \uElseIf{$m$ has the form \texttt{$x$\leadsfrom$m_1$;$m_2$}}{
      Call \cref{alg:is} on $m_1$ with the environment~\textit{env}, obtaining ($v_1$, $w_1$)\\
      Let $\textit{env}'$ be the environment \textit{env} extended with $x$ having the value $v_1$\\
      Call \cref{alg:is} on $m_2$ with the environment~$\textit{env}'$, obtaining ($v_2$, $w_2$)\\
      Return $(v_2, w_1 \cdot w_2)$}
    \Else{The other cases are similar to above}
 \caption{\small{Weighted sampler: sample($m$, \textit{env} = [])}}
 \label{alg:is}
\end{algorithm2e}

To make the semantics of our language more concrete, \cref{alg:is} shows
a sampler that takes a probabilistic program as input and returns a draw
from the distribution it represents.
It is our only operation that calls a random number
generator. We apply it last in a sequence of transformations
to perform approximate inference.

Like a typical interpreter, \cref{alg:is} takes as input not only
a program but also an environment, which is a table
mapping variable names to values.  Also, because our language includes
unnormalized measures, this sampler returns not only a draw but
also an importance weight.

\subsection{EXPECTATION TRANSFORMATION}
\label{s:expectation}

The rest of this section describes various inference transformations
that we apply to our probabilistic programs.  Because we implement some
transformations in terms of others, we describe the
transformations not in the order we apply them but in the order we
implement them.

Our expectation transformation turns any program that represents
a distribution into another program that represents its expected value.
This transformation is exact and simple even
though the expected values of many distributions have no closed form,
because our language represents integrals symbolically with \texttt{Int}.
For example, the expectation transformation turns the program
\texttt{x\,\leadsfrom\,Uniform(0,\,2);\ Uniform(x,\,3)}
into

\begin{BVerbatim}
Int(0,2,x, Int(x,3,y, y)/(3-x))/(2-0)
\end{BVerbatim}

The latter program represents the integral
\(
    \frac{1}{2-0} \int_0^2 \linebreak[-1] \frac{1}{3-x} \int_x^3 y \,dy \,dx \text.
\)
To compute this integral in closed form is to perform exact inference on
the given distribution.  The expectation transformation itself does not
do~so; nor does it approximate the integral by sampling.

\begin{algorithm2e}[t]
  \KwIn{program representing a measure: $m$}
  \KwIn{program representing a function: $f$}
  \KwOut{program representing a number}
\caption{\small{Expectation transformation: expect($m$, $f$)}}
  \textbf{Examine }$m$\\
  \uIf{$m$ has the form \texttt{Weight($w_1$,$e_1$)}}{ Return $w_1\cdot \texttt{App(}f\texttt{,}e_1\texttt{)}$ }
  \uElseIf{$m$ has the form \texttt{Normal($e_1$,$e_2$)}}{
     Return~\texttt{Int($-\infty$, $\infty$, x, $e_3 \cdot{}$App($f$,x))}
       where the program $e_3$ computes the density of the
       \texttt{Normal}($e_1$, $e_2$) distribution at \texttt{x}}
  \uElseIf{$m$ has the form \texttt{$x$\leadsfrom$m_1$;$m_2$}}{
     Call \cref{alg:expect} with $m_2$ and $f$ obtaining $e_3$\\
     Call \cref{alg:expect} with $m_1$ and \texttt{Lam($x$,$e_3$)}}
  \Else{The other cases are similar to above}
\label{alg:expect}
\end{algorithm2e}

Specified more generally, the expectation transformation turns any
program that represents a measure, along with a function from the sample space to
numbers%
% (such as the identity function)%
, into another program that
represents the
%(Lebesgue)
integral of the given function with respect to the
given measure.  We show this transformation as \cref{alg:expect}.  It handles
primitive distributions such as \texttt{Normal} by looking up their
density from a table.

\subsection{DENSITY AND DISINTEGRATION}
\label{ssec:disintegrate}

Turning a distribution into its density function is naturally expressed
as a program transformation \citep{bhat-type,BBGR13}.  More precisely,
the density transformation takes as input a probabilistic program
representing a distribution, and returns another program representing a
function that maps each point in the sample space to the density at that
point. Note that this transformation does not compute any density
numerically.  It only returns a program
that computes densities when interpreted by our weighted sampler
(\cref{alg:is}).
For example, the density transformation turns the probabilistic program

\begin{BVerbatim}
x <~ Uniform(0, 2);
y <~ Uniform(x, 3); Dirac((x, y))
\end{BVerbatim}

into the density function
\texttt{Lam((x,y),
If(0<x<2, If(x<y<3, 1/(3-x), 0)/(2-0), 0))}.

We implement density in terms of another program
transformation, \emph{disintegration}
\citep{shan-exact,narayanan-symbolic}.
Disintegration is similar to conditioning in that it
takes a probabilistic program representing a joint distribution
$\Pr(X,Y)$ as input, but instead of returning a conditional distribution
$\Pr(Y\mid X=x)$, disintegration returns an unnormalized slice
$\Pr(Y,X=x)$ of the original distribution.  More precisely,
disintegration returns a program representing a function from values
of~$x$ to measures $\Pr(Y,X=x)$.  Such a (measurable) function is also
known as a \emph{kernel}.

\begin{algorithm2e}[t]
  \KwIn{program representing a measure: $m$}
  \KwIn{program representing value drawn from $m$: $t$}
  \KwOut{program representing a nonnegative number}
   \begin{enumerate*}
     \item{Disintegrate \texttt{x\leadsfrom$m$; Dirac((x,Unit))},
       obtaining $e_1$}
     \item{Call \cref{alg:expect} on \texttt{App($e_1$,$t$)} and \texttt{Lam(y,1)}}
   \end{enumerate*}
   \unskip
\caption{Density transformation: density($m$, $t$)}
\label{alg:density}
\end{algorithm2e}

Taking advantage of the fact that disintegration does not normalize the
measures it returns,
we implement the density transformation in terms of disintegration and
expectation.  This implementation is shown in \cref{alg:density}.  It
invokes disintegration (letting $Y$ be the space that consists of a
single point~\texttt{Unit}) then expectation (letting the integrand~$f$
be the function that maps \texttt{Unit} to~$1$).

\begin{algorithm2e}[t]
  \KwIn{program representing a measure: $m$}
  \KwIn{program representing value drawn from $m$: $t$}
  \KwOut{program representing a measure}
  \textbf{Examine }$m$\\
  \uIf{$m$ has the form \texttt{Uniform($e_1$,$e_2$)} or
                        \texttt{Normal($e_1$,$e_2$)} or
                        \texttt{Gamma($e_1$,$e_2$)}}{
     Let $d$ be a program that computes the density of the
       distribution~$m$\\
     Return~\texttt{Weight(App($d$,$t$), $t$)}}
  \uElseIf{$m$ has the form \texttt{$x$\leadsfrom$m_1$;$m_2$}}{
     Call \cref{alg:observe} recursively with $m_2$ and $t$ obtaining $m_3$\\
     Return~\texttt{$x$\leadsfrom$m_1$;$m_3$}}
  \Else{Raise an error about not being able to handle $m$}
\caption{Observation transformation:~observe($m$, $t$)}
\label{alg:observe}
\end{algorithm2e}

Disintegration is useful independently of the density transformation.
For example, \cref{sec:example} uses it to
turn the prior \texttt{kalman} into the posterior
\texttt{kalman2}.

We sketch how disintegration works in terms of a simpler
program transformation, which we call \emph{observation}
(\cref{alg:observe}). This
transformation takes as input a measure~$m$ and a value~$t$ that could have
been drawn from~$m$, and returns a measure which only returns~$t$,
weighted by how likely that value was to be drawn from~$m$.
For example, the observation transformation turns the program

\begin{BVerbatim}
x <~ Uniform(0, 2); Uniform(x, 3)
\end{BVerbatim}

and the variable~\texttt{y} into the program

\begin{BVerbatim}
x <~ Uniform(0, 2);
Weight(If(x<y<3, 1/(3-x), 0), y)
\end{BVerbatim}

As indicated at the bottom in \cref{alg:observe}, the observation
transformation only handles a subset of our language. In particular, it
does not handle \texttt{Dirac}, so it does not handle the typical
program \texttt{kalman} in \cref{sec:example}.  In general, if the input
program performs arithmetic or any other deterministic computation to
produce the observation~$t$, then we need to invert this deterministic
computation and insert any Jacobian factors required. This inversion is
what the disintegration provides over observation.

To relate observation and disintegration more precisely, suppose the
program~$m$ represents a measure over~$X$, the program~$e$ represents a
value in~$Y$, and observation turns $m$ and~\texttt{x} into~$m_1$.  Then
disintegrating the program
\texttt{...; x\leadsfrom\(m\); Dirac((x,\(e\)))}
%(which uses \texttt{Dirac})
yields a program equivalent to
\texttt{Lam(x, ...; dummy\leadsfrom\(m_1\); Dirac(\(e\)))}.

\subsection{NORMALIZATION AND CONDITIONING}
\label{ssec:normalization}

The presence of \texttt{Weight} in our language enables the observation and
disintegration transformations to return measures that are typically
unnormalized.  To recover a probability distribution, we must reweight
the measure. We define this \emph{normalization} operation as a program
transformation as well, shown as \cref{alg:norm}.

\begin{algorithm2e}[t]
  \KwIn{program representing a measure: $m$}
  \KwOut{program representing a probability distribution}
   \begin{enumerate*}
    \item Call Algorithm~\ref{alg:expect} on~$m$ and
          \texttt{Lam(x,1)}\\ obtaining the program $e_1$
    \item Return \texttt{x\leadsfrom$m$; Weight(1/$e_1$, x)}
   \end{enumerate*}
   \unskip
\caption{\small{Normalization transformation: normalize($m$)}}
\label{alg:norm}
\end{algorithm2e}

\label{conditioning}
Conditioning can now be defined by composition:
it is just disintegration, followed by normalizing the measure.

\subsection{MCMC SAMPLING TRANSFORMATIONS}
\label{s:mh}

\begin{algorithm2e}[t]
  \KwIn{program representing the proposal kernel:~\textit{proposal}}
  \KwIn{program representing the target distribution:~\textit{target}}
  \KwOut{program representing MCMC transition kernel with acceptance ratio}
   \begin{enumerate*}
    \item Let \texttt{old} and \texttt{new} be fresh variable names
    \item{Call \cref{alg:density} on \textit{target} and \texttt{old}, obtaining
      $p_{\text{old}}$}
    \item{Call \cref{alg:density} on \textit{target} and \texttt{new}, obtaining
      $p_{\text{new}}$}
    \item{Call \cref{alg:density} on \texttt{App($\mathit{proposal}$,\texttt{new})} and \texttt{old},
      obtaining $q_\text{old;new}$}
    \item{Call \cref{alg:density} on \texttt{App($\mathit{proposal}$,\texttt{old})} and \texttt{new},
      obtaining $q_\text{new;old}$}
    \item{Let $e_1$ be $(p_{\text{new}} \cdot q_{\text{old;new}})/(p_{\text{old}} \cdot q_{\text{new;old}})$}
    \item{Return
        \texttt{\begin{tabular}[t]{@{}l@{}}
                Lam(old,\,new\leadsfrom App($\mathit{proposal}$,old);\\
                \ \ \ \ \ \ \ \ \,Dirac((new, $e_1$)))
                \end{tabular}}}
  \end{enumerate*}
  \unskip
\caption{Metropolis-Hastings sampling transformation: mh(\textit{proposal}, \textit{target})}
\label{alg:mh}
\end{algorithm2e}

A major contribution of this paper is to implement Markov chain Monte
Carlo (MCMC) methods, such as MH sampling and Gibbs
sampling, in a way that applies to a variety of target distributions and
composes with other inference techniques.  We express an
MCMC method as a transformation from a program representing the target
distribution to a program representing the transition kernel.  Whereas
the transformation itself makes no random choices, the latter program
can be interpreted by our weighted sampler (\cref{alg:is}) to generate a
random chain, or subject to simplification
(\cref{simpl}).

Following this approach, our MCMC implementations
closely resemble their textbook presentation.  As shown in
\cref{alg:mh}, where the textbook presentation of the acceptance ratio
refers to the target and proposal densities, our implementation invokes
the density transformation (\cref{alg:density}) on two probabilistic
programs, representing the target and proposal distributions.  Using
the fact that the density transformation symbolically
handles free variables such as \texttt{old} and \texttt{new}, we perform
the transformation just once (not once per sampler iteration) to
generate a program that takes the current state as input.

\begin{algorithm2e}[t]
  \KwIn{program representing the $n$-dimensional target distribution:~\textit{target}}
  \KwOut{program representing MCMC transition kernel}
    Let $\mathbf{x}$ be the set of the $n$ variables in the \textit{target}\\
    Initialize \textit{choices} to the empty sequence \texttt{[]}\\
    For each $x_i\in\mathbf{x}$:
      \begin{enumerate*}
        \item{Let $x_{-i}$ be the rest of the variables}
        \item{Let $e_1$ be
              \texttt{$\mathbf{x}$\leadsfrom\textit{target}; Dirac(($x_{-i}$,$x_i$))}}
        \item{Disintegrate $e_1$, obtaining $e_2$}
        \item{Let $e_3$ be \texttt{App($e_2$, $x_{-i}$)}}
        \item{Call \cref{alg:norm} on $e_3$, obtaining $e_4$}
        \item{Let $\mathbf{y}$ be $\mathbf{x}$ except replacing $x_i$ by \texttt{new}}
        \item{Let $e_5$ be \texttt{new\leadsfrom$e_4$; Dirac($\mathbf{y}$)}}
        \item{Add the pair \texttt{($1/n$, $e_5$)} to \textit{choices}}
      \end{enumerate*}
      \unskip
    Return \texttt{Lam($\mathbf{x}$, Superpose(\textrm{\textit{choices}}))}\\
\caption{Gibbs sampling transformation: gibbs(\textit{target})}
\label{alg:gibbs}
\end{algorithm2e}

Gibbs sampling is a special case of MH\@, where
the proposal kernel combines the results of conditioning the target
distribution along each dimension.  The acceptance
ratio is then always~1, so it need not be computed.
To produce such a proposal kernel automatically, we implement Gibbs
sampling as a separate transformation, \cref{alg:gibbs}.  The input is
a program representing an $n$\hyp dimensional distribution
$\Pr(x_1,\dotsc,x_n)$.  For each
random variable~$x_i$, we condition (\cref{conditioning})
on the other variables~$x_{-i}$ to get a
program that resamples~$x_i$.  We then combine these $n$
programs to form the proposal kernel.

\section{SIMPLIFICATION}
\label{simpl}

Because we express each inference technique as a transformation that
produces a probabilistic program, rather than as an
interpreter that makes immediate random choices, we can optimize
and simplify the produced programs.  To this end, we apply the
optimizations discussed by \citet{DBLP:conf/padl/CaretteS16}.  This
\emph{simplification} transformation does not change the measure
represented by a program but tries to place the program in a form that,
when interpreted by our weighted sampler (\cref{alg:is}),
draws samples faster and with more uniform weights.

Based on computer algebra, the simplification transformation recognizes
conjugacy relationships, integrates out latent variables, and performs
algebraic simplifications.
Like other transformations, simplification operates on a program
before any variables receive their values, so in particular its efficacy
is independent of data sizes.
The rest of this section briefly describes
these optimizations.

\paragraph{Conjugacy relationships}

Simplification recognizes when a density represented
by \texttt{Weight} matches the density of a primitive distribution.
A simple example arises from the joint distribution $\Pr(Y,X)$
represented below:

\begin{BVerbatim}
x <~ Normal(a, s);
y <~ Normal(x, t); Dirac((y, x))
\end{BVerbatim}

Disintegrating this program (\cref{ssec:disintegrate})
produces

\begin{BVerbatim}
x <~ Normal(a, s);
Weight( exp(-(y-x)^2/(2*t^2))
        /t/sqrt(2*pi)         , x )
\end{BVerbatim}

This latter program scales the measure \texttt{Normal(a,s)} with
the density \verb|exp(...)/t/sqrt(2*pi)| to
represent the conditional distribution $\Pr(X\mid Y)$ up to a constant
factor.  Normalizing and simplifying it yields

\begin{BVerbatim}
Normal( (y*s^2+a*t^2)/(s^2+t^2),
        s*t/sqrt(s^2+t^2) )
\end{BVerbatim}

using the conjugacy relationship between \texttt{Normal}s
(assuming \texttt{s} and~\texttt{t} are positive).
This simplified program runs faster; it~draws samples
without weighting them.

This optimization is symbolic, in the sense that it works even when the
initial program contains free variables such as \texttt{a}, \texttt{s},
and~\texttt{t}, whose values are unknown.

This optimization is robust because it recognizes not
words like \texttt{Normal} but the densities they denote.  Thus it works
even if we express \texttt{Normal(0,1)} by spelling out its density,
whether we expand the polynomial \verb|-(y-x)^2|.
All conjugacies among \texttt{Normal}, \texttt{Gamma}, and
\texttt{Beta} thus fall out from recognizing their densities.

\paragraph{Integrating out a variable}

When a distribution is described using a latent random variable, it
usually helps to eliminate the variable.
Such latent variables include \texttt{x1}
and \texttt{x2} in \cref{sec:example}, as well as \texttt{x} in

\begin{BVerbatim}
x <~ Normal(0, 1); Normal(x, 1)
\end{BVerbatim}

The simplification transformation eliminates these variables.  In
particular, it integrates out continuous latent
variables symbolically.  The density-recognition machinery just described
then produces simpler, faster, and equivalent programs,
such as
\texttt{Normal(0,sqrt(2))}.

This integration is symbolic, again in the sense that it works even when
the initial program contains free variables whose values are unknown.
For example, the program
\verb|x <~ Normal(a, s); Normal(x, t)|
simplifies to \verb|Normal(a, sqrt(s^2+t^2))|.

\paragraph{Algebraic simplifications}

When we produce a program that calculates acceptance ratios, the
numerator and denominator share many factors, which are usually
canceled out by hand.  The simplification
transformation automates this optimization using computer algebra, so
an expression like \texttt{($a$*$b$)/($a$*$c$)} becomes
\texttt{$b$/$c$}.

\section{EXPERIMENTAL RESULTS}
\label{experiments}

To demonstrate that our approach is modular and practical, we apply
multiple inference methods (MH, Gibbs) to a variety of models.
We conduct three
experiments using the Hakaru system \citep{flops/Hakaru2016}.

\emph{Modular} means we can re-use
the components in \cref{transforms} and \cref{simpl} to produce all
three samplers. In each experiment, a pipeline composed of reusable inference
transformations turns a concise generative model into an executable
MCMC sampler in seconds.

\emph{Practical} means our approach
can solve real-world problems by expressing popular models and
inference methods discussed in the literature. The largest of our
three experiments is the third, a document classification task using
the 20 Newsgroups corpus.

We measure the accuracy and speed of our automatically generated
samplers, showing they are in line with solutions from commonly used
probabilistic programming languages. Our samplers are more accurate
across the board because simplification eliminates all latent
continuous variables, regardless of the dimensionality of the problem
(that is, input and output array sizes).

All measurements were produced on a quad-core Intel i5-2540M
processor running 64-bit Ubuntu~16.04\@. Our samplers use Glasgow
Haskell Compiler 8.0.1 \texttt{-O2}.

\subsection{MH SAMPLING FOR DYNAMICS}
\label{easyroad}

\begin{table}
\centering
\caption{MH sampler run times for linear dynamics}
\label{fig:kalmanspeed}
\begin{tabular}{ l r r }
\toprule
  \textbf{Inference method} & & \llap{\textbf{Run time} (msecs)} \\
  & \llap{Mean} & SD \\
\midrule
  WebPPL & 1078 & 16 \\
  Hakaru without simplifications & 1321 & 93 \\
  Hakaru with simplifications & 269 & 10 \\
  Handwritten & 207 & 4 \\
\bottomrule
\end{tabular}

\vskip\floatsep
\caption{MH sampler ESS rates for linear dynamics}
\label{fig:kalmanacc}
\begin{tabular}{ l r r }
\toprule
\textbf{Inference method} & \multicolumn{2}{c}{\textbf{ESS per sample}} \\
    & $\mathrm{noise}_T$ & $\mathrm{noise}_E$ \\
\midrule
  WebPPL & 0.03 & 0.01 \\
  Hakaru & 0.09 & 0.34 \\
\bottomrule
\end{tabular}
\end{table}

In our first
experiment, we use MH to sample the random parameters of the linear dynamical system
in \cref{sec:example}.
We compare our generated samplers with one produced by WebPPL\@,
a state-of-the-art probabilistic programming system, and with
one written by hand.
The WebPPL sampler was compiled to JavaScript using Node 0.10.28\@.

\Cref{fig:kalmanspeed} shows that our system generates a fast sampler,
measured by using each sampler to draw 20,000 samples 10 times.
Thanks to the simplifications that turn \texttt{kalman4} into
\texttt{kalman5}, the Hakaru sampler is 4 times as fast as the WebPPL
sampler for the conditional distribution \texttt{kalman2}.
(These times exclude
the few seconds each system takes to compile the
model into a sampler.)

\Cref{fig:kalmanacc} shows that our samplers generate good samples,
quantified by the Effective Sample Size (ESS\@).
Our ESS is higher per sample
compared to WebPPL\@, because latent variables have been
integrated out in \texttt{kalman3}.

%% \remark{For the purpose of supporting the first sentence of this
%% paragraph, the variance measurement seems uninformative, so perhaps omit
%% it?}
%% We also measure the
%% variance (over 10 runs) of the mean (of 1000 samples per run) for each
%% sampler: for WebPPL\@, the variance is 0.007 for \texttt{noiseT} and
%% 0.009 for \texttt{noiseE}\@; for us, the variance is 0.009 for
%% \texttt{noiseT} and 0.002 for \texttt{noiseE}\@.

\subsection{GIBBS SAMPLING FOR CLASSIFICATION}

%% The second experiment is to infer topics from a small latent Dirichlet
%% allocation model.  The model only has 2 documents, 3 words, and 2
%% latent topics.  In \cref{fig:ldaspeed}, we compare the run times of
%% three Gibbs samplers, the first two of which were produced
%% automatically using our program transformations. For the first row, we
%% attempt to run the output of \cref{alg:gibbs} directly. Because that
%% output contains \texttt{Int}, we must perform numerical integration,
%% which takes a long time. For the second row, we feed that output of
%% \cref{alg:gibbs} through simplification, thereby avoiding numerical
%% integration. For the last row, we use a Gibbs sampler handwritten to
%% take advantage of conjugacies.

In our second and third experiments, we generate
Gibbs samplers and compare them to JAGS
(v4.20\@), a probabilistic programming system widely considered practical
for Gibbs sampling.  We measure accuracy by how well the samplers
recover true classifications, and speed by
the time it takes to produce samples. This
time consists of \emph{initialization} time and time spent
actually sampling.
Initialization time is the time a system takes from receiving the model
to generating the first sample: for JAGS
to load the model into memory, and for Hakaru
to simplify the model and compile the result into machine code.

Our second experiment is to classify synthetic data using a
Gaussian mixture model that has 3 components.

\begin{figure}
\centering
\includegraphics[width=0.4\textwidth]{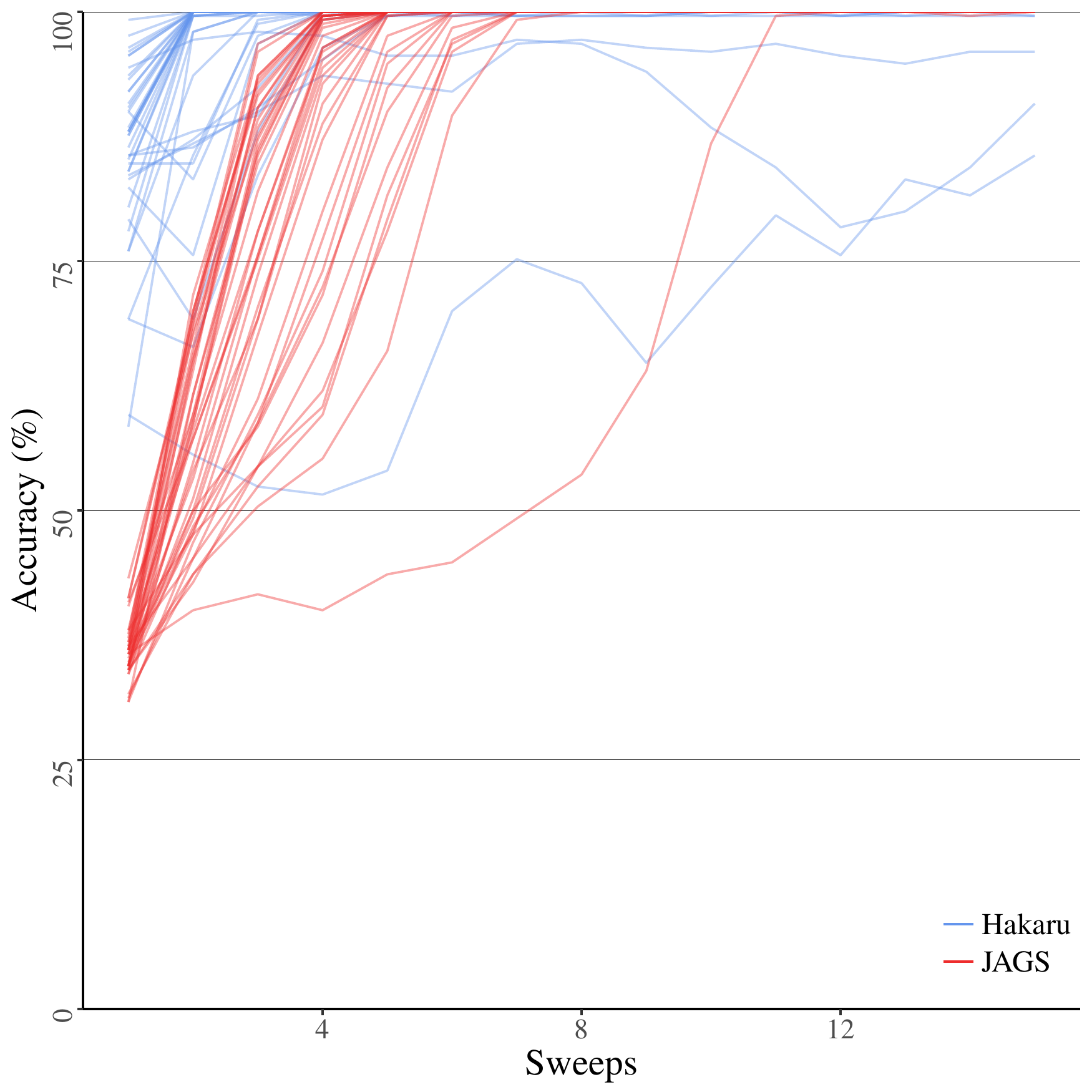}
\vspace{-8pt}
\caption{Gibbs sampler accuracy for Gaussian mixture}
\label{fig:gmmsweeps}

\vskip\floatsep
\includegraphics[width=0.4\textwidth]{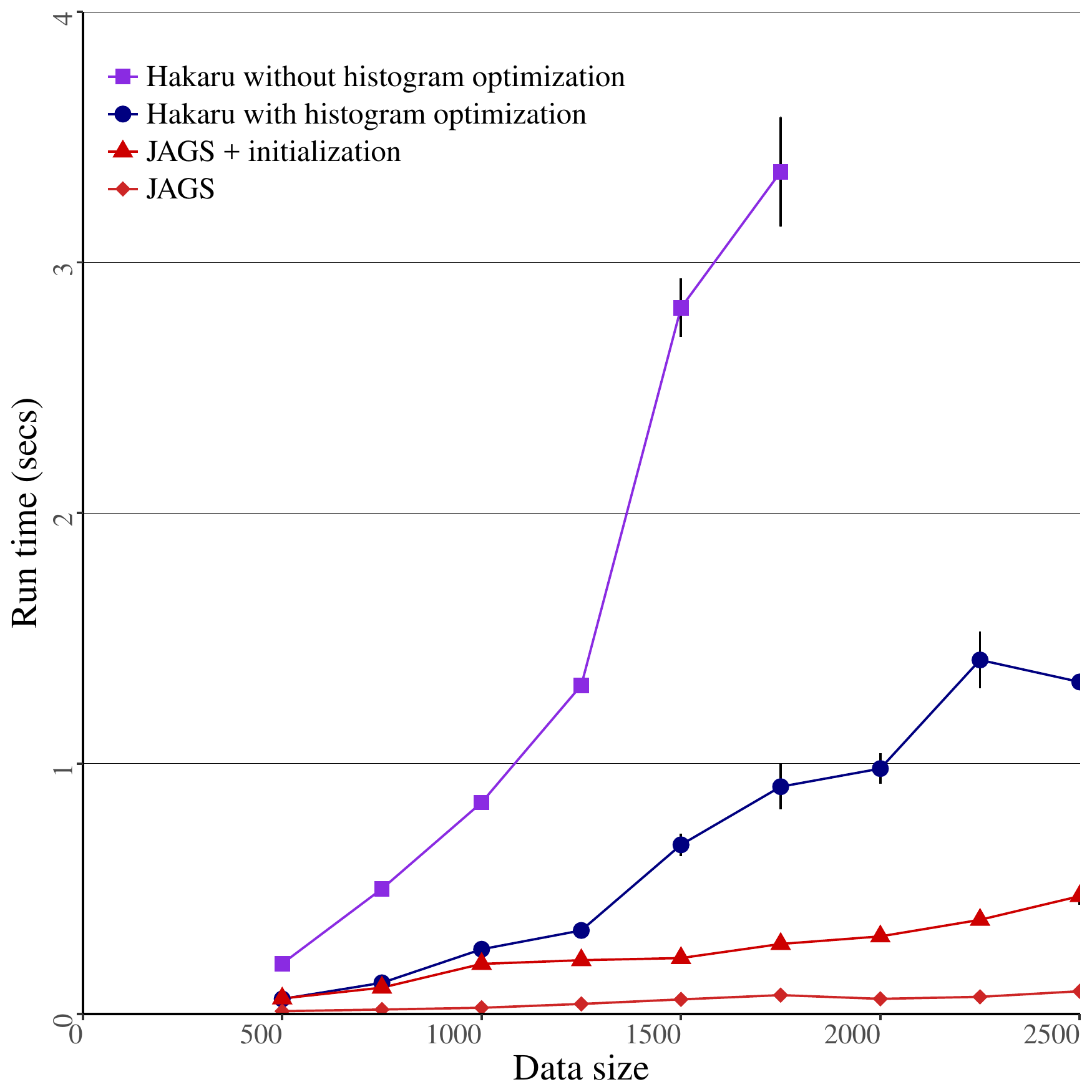}
\vspace{-8pt}
\caption{Gibbs sampler run times for Gaussian mixture}
\label{fig:gmmspeed}
\end{figure}

\Cref{fig:gmmsweeps} shows that Hakaru requires fewer sweeps than JAGS
to achieve the same accuracy.  Each curve plots the accuracy of one
chain over the course of 15 sweeps on 250 data points.  After just one
sweep, all 20 Hakaru chains are $>50\%$ accurate, unlike the 20 JAGS
chains, which take a few sweeps to catch up.  The cause is that Hakaru's
simplification transformation recovers a collapsed Gibbs sampler that
computes the sample mean and variance of each mixture component in
closed form.

\Cref{fig:gmmspeed} shows Hakaru is about one order of magnitude slower
than JAGS\@, measured by how long 6 sweeps take, varying data size
from 500 to 2500 points. The top
two curves represent two samplers generated by Hakaru with different
lower-level optimizations:
% both
% use common-subexpression elimination and
% loop-invariant code motion, but
the second-from-top curve adds
a \emph{histogram} optimization to compute
summary statistics such as the per-mixture-component sum \( \sum_{i=1}^N
\bigl\{\begin{smallmatrix} t_i \hfill & \text{if $z_i=z^*$} \hfill
\\ 0 \hfill & \text{otherwise} \hfill \end{smallmatrix} \) in a single
pass over the data for all components~$z^*$.  The bottom two
curves show the run time of JAGS\@, with and
without initialization.
JAGS's speed advantage
can be explained by Hakaru's current inability to reuse
computation between updates during a sweep. Still,
Hakaru is practical for this real-world task.

\begin{figure}
\centering
\includegraphics[width=0.4\textwidth]{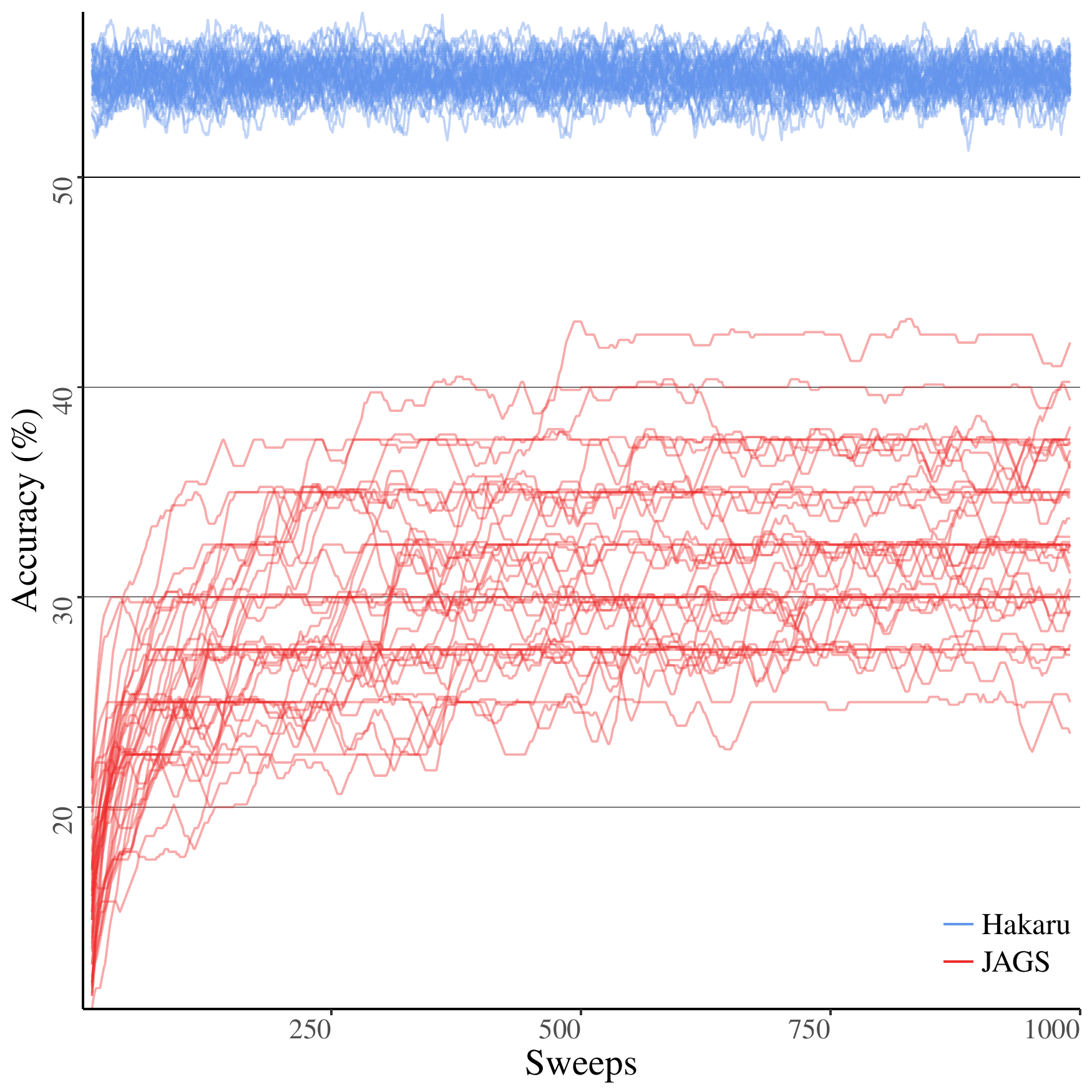}
\vspace{-8pt}
\caption{Document classification accuracy by
  sweeps}
\label{fig:nbsweeps}

\vskip\floatsep
\includegraphics[width=0.4\textwidth]{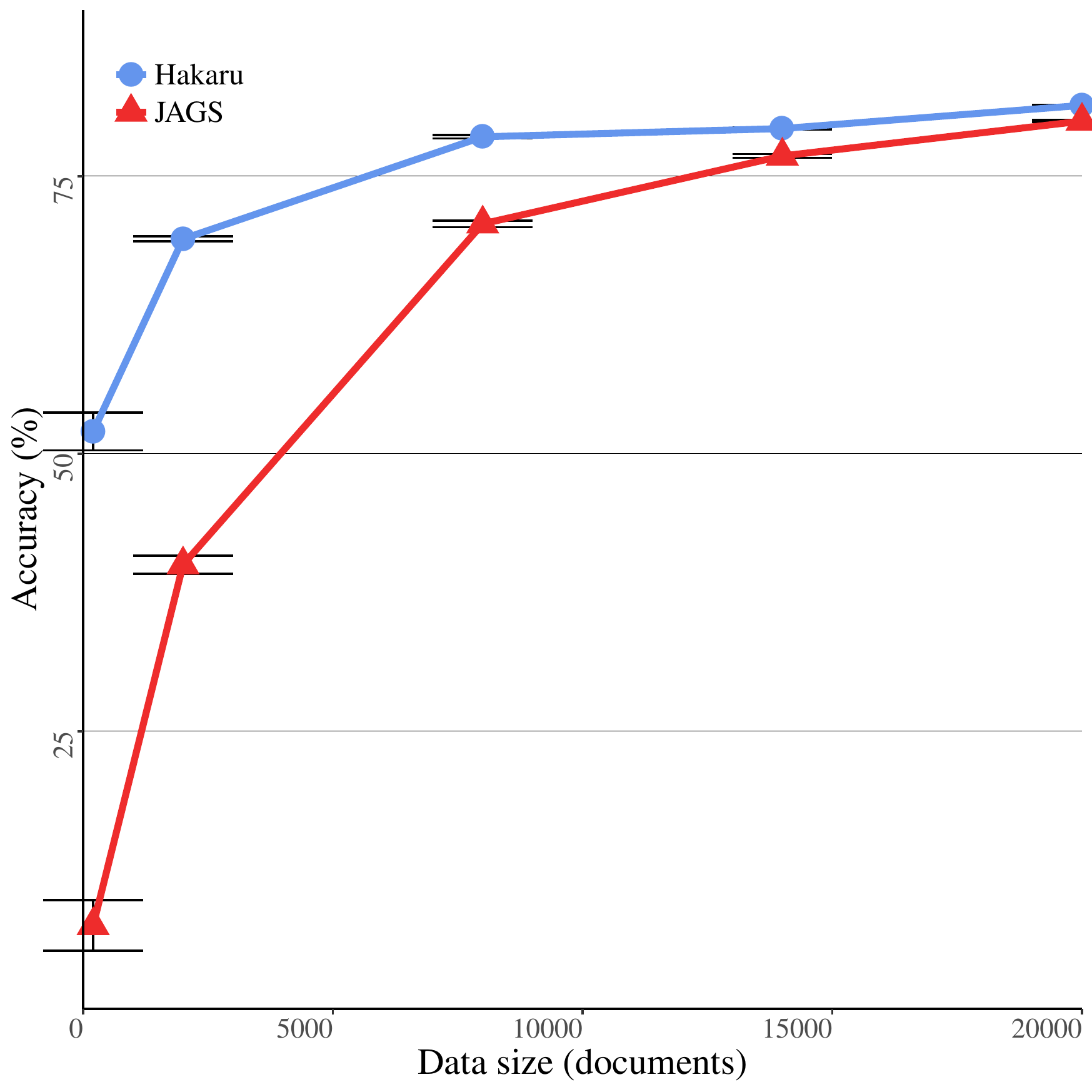}
\vspace{-8pt}
\caption{Document classification accuracy by
  data size}
\label{fig:nbaccuracy}

\vskip\floatsep
\includegraphics[width=0.4\textwidth]{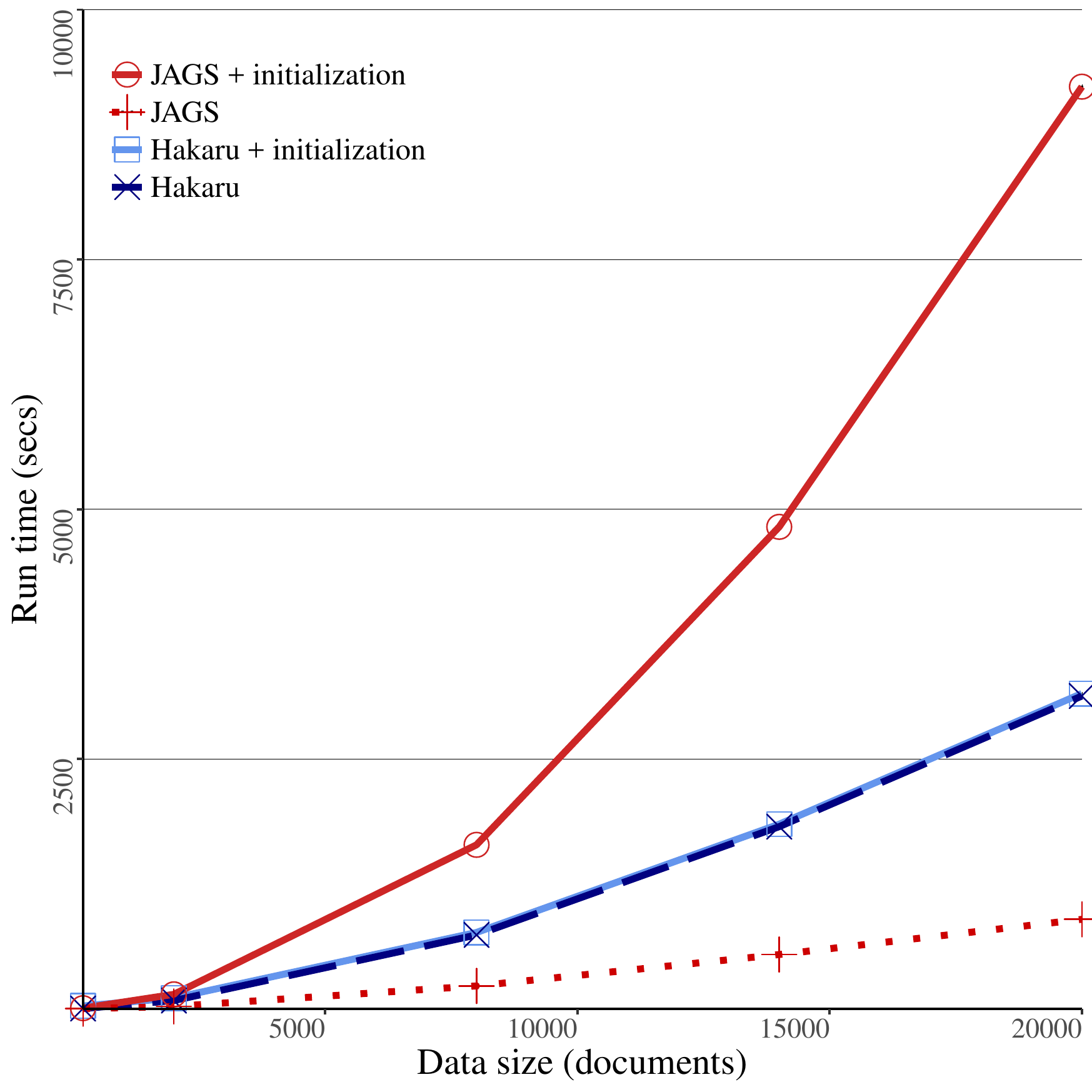}
\vspace{-8pt}
\caption{Document classification run times. Error bars were too small to display.}
\label{fig:nbspeed}
\end{figure}

In our third experiment, Hakaru generates a classifier for the 20
Newsgroups corpus that is more accurate than JAGS and comparable in
speed.  We use the same Multinomial Naive Bayes model and 20 Newsgroups
corpus as \citet{mccallum1998comparison}.  We hold out 10\% of the
labels and use Gibbs sampling to infer them.  We evaluate the samplers
on data sizes ranging from 200 to all 19997 documents, evenly
distributed among newsgroups.

Because the sampler generated by Hakaru is collapsed, it is more
accurate than JAGS in two ways. First, \cref{fig:nbsweeps} shows Hakaru
achieves better accuracy than JAGS after one sweep, and continues to for
at least 1000 sweeps. Each curve there plots the accuracy (moving average
with window size 20) of one chain on 400 documents.
Second, \cref{fig:nbaccuracy} shows Hakaru more accurate than JAGS
across data sizes.  We use 2 sweeps there since JAGS does not perform
above chance with only 1 sweep.

\Cref{fig:nbspeed} shows Hakaru is as fast as JAGS\@, measured by how
long 2 sweeps take, varying data size.  JAGS's initialization time grows
with the data size, while Hakaru's is constant.  Whereas JAGS unrolls
loops into a pointer-based stochastic graph whose size grows with the
data, Hakaru generates tight loops over unboxed arrays irrespective of
the data size.  Even disregarding initialization time, JAGS is at best
4 times faster than Hakaru.

%% We repeat the experiment 10 times to produce the mean and
%% standard deviations shown in \cref{fig:kalmanspeed,fig:gmmspeed}.

\section{CONCLUSIONS}

We express inference methods by composing program
transformations such as disintegration and expectation.  The resulting
modular inference procedures perform comparably to other
probabilistic programming systems and are usable for practical
problems. This technique makes it easier and faster to create and test
inference procedures and to explore novel inference methods.

\begin{comment}
Future work includes implementing single-site MH as a program
transformation that uses the dependency structure of a target distribution
to generate a proposal kernel automatically. We
also want to incorporate Hamiltonian Monte Carlo and variational
inference, which call for automatic differentiation as a program
transformation.
Finally, we are implementing slice sampling as a program transformations
whose definition resembles its textbook presentation.
\end{comment}

\subsubsection*{Acknowledgements}
We thank David Belanger, Jacques Carette, and Chad Scherrer for helpful discussions,
feedback, and suggestions. This research was supported by DARPA contract
FA8750-14-2-0007, NSF grant CNS-0723054, Lilly Endowment,
Inc., and the Indiana METACyt Initiative.

\bibliographystyle{plainnat}
\bibliography{multi}

\begin{thebibliography}{19}
\providecommand{\natexlab}[1]{#1}
\providecommand{\url}[1]{\texttt{#1}}
\expandafter\ifx\csname urlstyle\endcsname\relax
  \providecommand{\doi}[1]{doi: #1}\else
  \providecommand{\doi}{doi: \begingroup \urlstyle{rm}\Url}\fi

\bibitem[Bhat et~al.(2012)Bhat, Agarwal, Vuduc, and Gray]{bhat-type}
Sooraj Bhat, Ashish Agarwal, Richard Vuduc, and Alexander Gray.
\newblock A type theory for probability density functions.
\newblock In \emph{{POPL} '12: Conference Record of the Annual {ACM} Symposium
  on Principles of Programming Languages}, pages 545--556. {ACM} {P}ress,
  January 2012.

\bibitem[Bhat et~al.(2013)Bhat, Borgstr{\"o}m, Gordon, and Russo]{BBGR13}
Sooraj Bhat, Johannes Borgstr{\"o}m, Andrew~D. Gordon, and Claudio Russo.
\newblock Deriving probability density functions from probabilistic functional
  programs.
\newblock In \emph{19th International Conference on Tools and Algorithms for
  the Construction and Analysis of Systems ({TACAS})}, 2013.

\bibitem[Carette and Shan(2016)]{DBLP:conf/padl/CaretteS16}
Jacques Carette and Chung{-}chieh Shan.
\newblock Simplifying probabilistic programs using computer algebra.
\newblock In \emph{Practical Aspects of Declarative Languages - 18th
  International Symposium, {PADL} 2016, St. Petersburg, FL, USA, January 18-19,
  2016. Proceedings}, pages 135--152, 2016.

\bibitem[{Gelman} et~al.(2014){Gelman}, {Vehtari}, {Jyl{\"a}nki}, {Robert},
  {Chopin}, and {Cunningham}]{2014arXiv1412.4869G}
A.~{Gelman}, A.~{Vehtari}, P.~{Jyl{\"a}nki}, C.~{Robert}, N.~{Chopin}, and
  J.~P. {Cunningham}.
\newblock {Expectation Propagation as a Way of Life}.
\newblock \emph{ArXiv e-prints}, December 2014.

\bibitem[Goodman et~al.(2008)Goodman, Mansinghka, Roy, Bonawitz, and
  Tenenbaum]{GMR+08}
Noah~D. Goodman, Vikash~K. Mansinghka, Daniel~M. Roy, Keith Bonawitz, and
  Joshua~B. Tenenbaum.
\newblock Church: a language for generative models.
\newblock In \emph{Proc. of Uncertainty in Artificial Intelligence}, 2008.

\bibitem[Griffiths and Steyvers(2004)]{griffiths2004finding}
Thomas~L Griffiths and Mark Steyvers.
\newblock Finding scientific topics.
\newblock \emph{Proceedings of the National Academy of Sciences}, 101\penalty0
  (suppl 1):\penalty0 5228--5235, 2004.

\bibitem[Hughes et~al.(2015)Hughes, Kim, and Sudderth]{hughes2015reliable}
Michael Hughes, Dae~Il Kim, and Erik Sudderth.
\newblock Reliable and scalable variational inference for the hierarchical
  {D}irichlet process.
\newblock In \emph{Proceedings of the Eighteenth International Conference on
  Artificial Intelligence and Statistics}, pages 370--378, 2015.

\bibitem[McCallum and Nigam(1998)]{mccallum1998comparison}
Andrew McCallum and Kamal Nigam.
\newblock A comparison of event models for naive {B}ayes text classification.
\newblock In \emph{AAAI-98 workshop on learning for text categorization},
  volume 752, pages 41--48. Citeseer, 1998.

\bibitem[McCallum et~al.(2008)McCallum, Rohanemanesh, Wick, Schultz, and
  Singh]{MRWSS08}
Andrew McCallum, Khashayar Rohanemanesh, Michael Wick, Karl Schultz, and Sameer
  Singh.
\newblock Factorie: Efficient probabilistic programming for relational factor
  graphs via imperative declarations of structure, inference and learning.
\newblock In \emph{NIPS Workshop on Probabilistic Programming}, 2008.

\bibitem[Narayanan and Shan(2017)]{narayanan-symbolic}
Praveen Narayanan and Chung-chieh Shan.
\newblock Symbolic conditioning of arrays in probabilistic programs.
\newblock In \emph{{ICFP} '17: Proceedings of the {ACM} International
  Conference on Functional Programming}. {ACM} {P}ress, 2017.

\bibitem[Narayanan et~al.(2016)Narayanan, Carette, Romano, Shan, and
  Zinkov]{flops/Hakaru2016}
Praveen Narayanan, Jacques Carette, Wren Romano, Chung-chieh Shan, and Robert
  Zinkov.
\newblock Probabilistic inference by program transformation in {H}akaru (system
  description).
\newblock In Oleg Kiselyov and Andy King, editors, \emph{Proceedings of {FLOPS}
  2016: 13th International Symposium on Functional and Logic Programming},
  number 9613 in {L}ecture {N}otes in {C}omputer {S}cience, pages 62--79.
  Springer, 2016.

\bibitem[Neiswanger et~al.(2014)Neiswanger, Wang, and
  Xing]{neiswanger2014asymptotically}
Willie Neiswanger, Chong Wang, and Eric Xing.
\newblock Asymptotically exact, embarrassingly parallel {MCMC}.
\newblock In \emph{The Conference on Uncertainty in Artificial Intelligence
  (UAI)}, 2014.

\bibitem[Pfeffer(2009)]{pfeffer2009figaro}
Avi Pfeffer.
\newblock {F}igaro: An object-oriented probabilistic programming language.
\newblock \emph{Charles River Analytics Technical Report}, 137, 2009.

\bibitem[Riedel et~al.(2014)Riedel, Singh, Srikumar, Rockt{\"a}schel,
  Visengeriyeva, and Noessner]{RSSRVN2014}
Sebastian Riedel, Sameer Singh, Vivek Srikumar, Tim Rockt{\"a}schel, Larysa
  Visengeriyeva, and Jan Noessner.
\newblock {WOLFE: Strength Reduction and Approximate Programming for
  Probabilistic Programming}.
\newblock In \emph{International Workshop on Statistical Relational AI
  (StarAI)}, 2014.

\bibitem[{\'S}cibior and Ghahramani(2016)]{scibiormodular}
Adam {\'S}cibior and Zoubin Ghahramani.
\newblock Modular construction of {B}ayesian inference algorithms.
\newblock In \emph{NIPS Workshop on Advances in Approximate Bayesian
  Inference}, 2016.

\bibitem[{\'S}cibior et~al.(2015){\'S}cibior, Ghahramani, and
  Gordon]{DBLP:conf/haskell/ScibiorGG15}
Adam {\'S}cibior, Zoubin Ghahramani, and Andrew~D. Gordon.
\newblock Practical probabilistic programming with monads.
\newblock In \emph{Proceedings of the 8th {ACM} {SIGPLAN} Symposium on Haskell,
  Haskell 2015, Vancouver, BC, Canada, September 3-4, 2015}, pages 165--176,
  2015.

\bibitem[Shan and Ramsey(2017)]{shan-exact}
Chung-chieh Shan and Norman Ramsey.
\newblock Exact {B}ayesian inference by symbolic disintegration.
\newblock In \emph{{POPL} '17: Conference Record of the Annual {ACM} Symposium
  on Principles of Programming Languages}, pages 130--144. {ACM} {P}ress, 2017.

\bibitem[Wood et~al.(2014)Wood, van~de Meent, and
  Mansinghka]{wood-aistats-2014}
Frank Wood, Jan~Willem van~de Meent, and Vikash Mansinghka.
\newblock A new approach to probabilistic programming inference.
\newblock In \emph{Proceedings of the 17th International conference on
  Artificial Intelligence and Statistics}, pages 1024--1032, 2014.

\bibitem[Xu et~al.(2014)Xu, Lakshminarayanan, Teh, Zhu, and
  Zhang]{NIPS20145596}
Minjie Xu, Balaji Lakshminarayanan, Yee~Whye Teh, Jun Zhu, and Bo~Zhang.
\newblock Distributed {B}ayesian posterior sampling via moment sharing.
\newblock In Z.~Ghahramani, M.~Welling, C.~Cortes, N.D. Lawrence, and K.Q.
  Weinberger, editors, \emph{Advances in Neural Information Processing Systems
  27}, pages 3356--3364, 2014.

\end{thebibliography}

\end{document}